\documentclass[runningheads]{llncs}
\usepackage[T1]{fontenc}
\usepackage{color, soul}
\usepackage{subfigure}
\usepackage{amsmath} 
\usepackage{booktabs}
\usepackage{multirow}

\usepackage{graphicx}
\usepackage{bbm}
\usepackage{amssymb} 
\usepackage{hyperref}
\begin{document}
%

\title{Enhancing Retinal Disease Classification from OCTA Images via Active Learning Techniques}
%
%
\author{Jacob Thrasher \inst{1} \and Annahita Amireskandari \inst{2} \and Prashnna Gyawali \inst{1}}

%
\institute{Lane Department of Computer Science and Electrical Engineering, West Virginia University \and
School of Medicine, West Virginia University}
%
\maketitle              
\begin{abstract}
Eye diseases are common in older Americans and can lead to decreased vision and blindness. Recent advancements in imaging technologies allow clinicians to capture high-quality images of the retinal blood vessels via Optical Coherence Tomography Angiography (OCTA), which contain vital information for diagnosing these diseases and expediting preventative measures. OCTA provides detailed vascular imaging as compared to the solely structural information obtained by common OCT imaging.
Although there have been considerable studies on OCT imaging, there have been limited to no studies exploring the role of artificial intelligence (AI) and machine learning (ML) approaches for predictive modeling with OCTA images. In this paper, we explore the use of deep learning to identify eye disease in OCTA images. However, due to the lack of labeled data, the straightforward application of deep learning doesn't necessarily yield good generalization. To this end, we utilize active learning to select the most valuable subset of data to train our model. We demonstrate that active learning subset selection greatly outperforms other strategies, such as inverse frequency class weighting, random undersampling, and oversampling, by up to 49\% in F1 evaluation. The full code can be found here: \href{https://github.com/jacob-thrasher/AL-OCTA}{https://github.com/jacob-thrasher/AL-OCTA}

\end{abstract}
\section{Introduction}

Eye diseases such as Age-related Macular Degeneration (AMD) and Diabetic Retinopathy (DR) can lead blurred, obstructed, or even total loss of vision \cite{eye,a2024_choroidal}. The American Academy of Ophthalmology reports that approximately 3 in 10 Americans aged 80 and older exhibit symptoms of AMD, and 1 in 4 Americans over 40 have DR \cite{a2015_eye}. Although it is not possible to reverse all of the damage caused by these diseases, early detection can prevent further deterioration \cite{eye}. Notably, it is estimated that one billion people globally suffer from sight limiting eye diseases that could have been prevented with proper treatment \cite{fricke_2018_global}. Therefore, early detection is crucial to prolonging the patient's eyesight. 

To understand eye physiology, various imaging technologies like retinal fundus photography and optical coherence tomography (OCT) are used. OCT, for example, employs light waves to create cross-sectional images of tissues, including the retina, and is commonly used to diagnose eye diseases. \textbf{Fig 1. (Top)} shows an example of OCT, which functions similarly to ultrasound but uses light instead of sound, providing 10-100 times finer images \cite{fujimoto_2000_optical}. These detailed images enable advancements in deep learning for identifying eye diseases via artificial intelligence. Task-specific models trained on OCT images can classify retinal macular \cite{rasti_2018_macular,wang_2019_on} and neuro-degenerative diseases, requiring a large corpus of annotated data for state-of-the-art results. Similarly, there has been advancement in foundation model capable of identifying ocular diseases like AMD, DR, and glaucoma, as well as systemic conditions such as Parkinson's, stroke, and heart failure \cite{zhou_2023_a}.

Optical coherence tomography angiography (OCTA), depicted in \textbf{Fig. 1 (Bottom)} is an advanced imaging technique that visualizes blood flow within the retina and choroid, unlike traditional OCT, which only captures structural images. 
OCTA aggregates multiple images to detect blood vessels in the eye via flow patterns, known as angiography. Other forms of angiography such as fluorescein and indocyanine-green require dye injections, which are invasive and considerably slower than OCTA \cite{spaide2018optical,sousa2020optical}. Additionally, the ability to capture blood vessel information of the choroid allows OCTA to detect other diseases such as AMD, choroidal neovascularization (CNV) and DR. Despite its potential, there are limited deep-learning studies focused on OCTA due to its complexity and the need for extensive annotated datasets.

In this paper, we investigate the capability of deep learning models to identify retinal diseases using an available OCTA dataset. Unfortunately, the publicly available datasets are limited in number and, more importantly, exhibit imbalance concerning diseased classes. To address this, we focus into data engineering aspects to ensure deep learning methods are applicable to this advanced imaging technique. Specifically, we employ active learning strategies to enhance the generalization of deep learning models.

We utilize a standard convolutional neural network and the OCTA500 dataset \cite{li2024octa} to construct our deep learning framework, exploring various data engineering strategies, from augmentation to active learning. Notably, active learning strategies demonstrate superior performance. Within the realm of active learning, we analyze the impact of instance sampling and subject sampling, as well as study the role of calibrating the network within the active learning framework. Our findings underscore the importance of these strategies in improving model performance on imbalanced OCTA datasets.


\begin{figure}[t]
    \centering
    \includegraphics[scale=0.28]{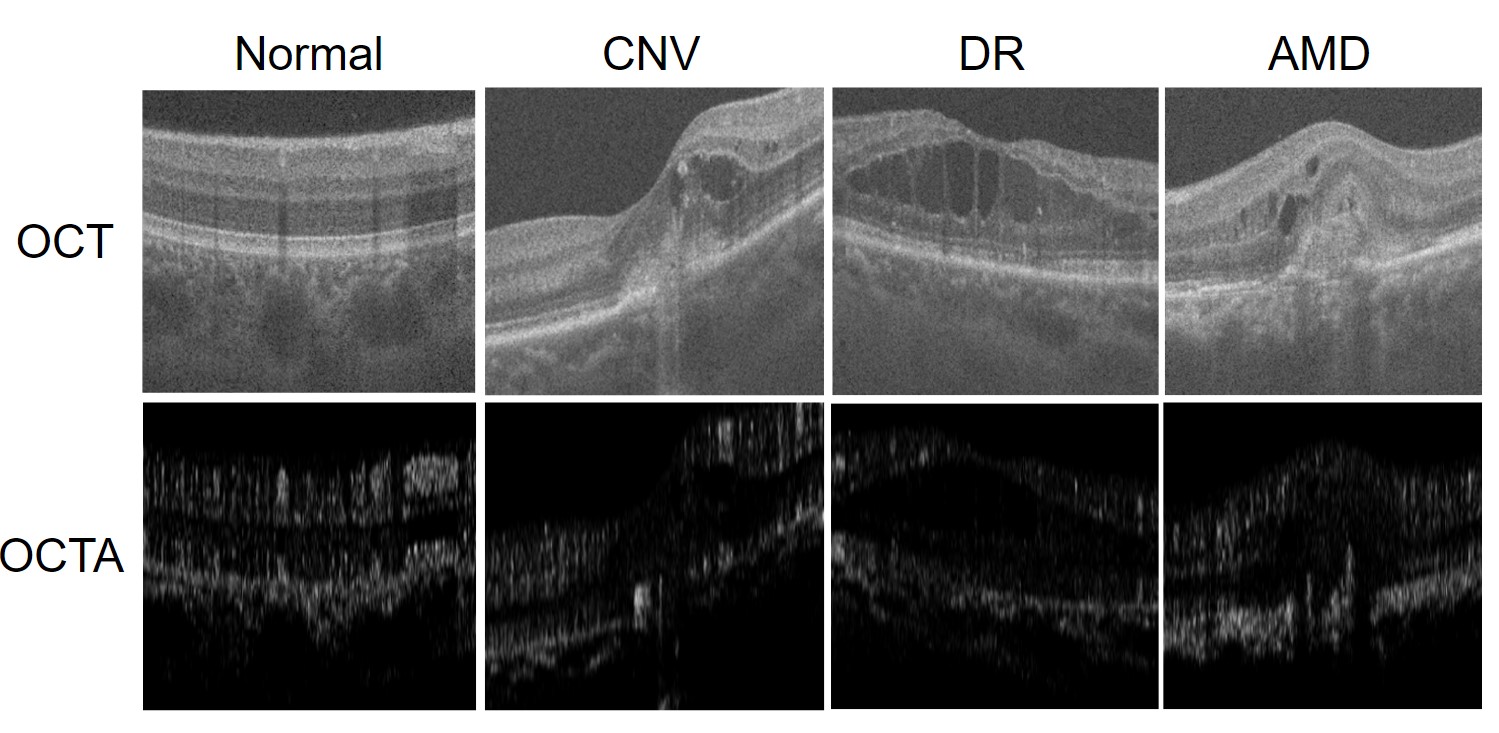}
    \caption{Comparison of OCT and OCTA data for Normal, CNV, DR, and AMD eyes}
    \label{fig:OCT}
\end{figure}


\section{Methods}

\subsection{Dataset}
We utilize the OCTA500 dataset \cite{li2024octa} for our experiments. This dataset contains 3mm $\times$ 3mm OCTA imaging samples for 200 subjects, each with 304 eye scans, yielding a total of 60800 OCTA images. Additionally, OCTA500 is a highly imbalanced dataset, containing 160 healthy retinas, 5 CNV, 6 AMD, and 29 DR. This imbalance demonstrates a strong need to employ data engineering techniques to adequately train a neural network. There is additionally a wide range of severity in each category, where some retinas exhibit minor abnormalities while others are more severely damaged. Lastly, there are some cases in this dataset which contain some overlap between diseases where, for example, CNV occurs due to AMD. OCTA500 does not provide detailed distinctions for these particular images, so we simply use the single labels that are included.

\subsection{Active learning}
Active learning (AL) is conventionally applied on a large corpus of unlabeled data, where human annotation is far too time consuming and costly to manually label every data point. Instead, a small subset of labeled data is first used to train a machine learning model. Then, the trained model attempts to classify the remaining unlabeled data elements and returns an uncertainty score for each, which measures the model's confidence for each prediction. With this, we can infer that predictions with high uncertainty represent difficult to classify data points, allowing developers to focus their labeling efforts on those instances. 

The same intuition can be applied to fully labeled datasets to determine which elements should be selected during undersampling. By choosing the instances which the model struggles with most, we can curate an optimal and balanced subset of our data for training. To accomplish this, we begin by training a baseline model on a perfectly balanced subset of 8 subjects (2 representing each class). Then, each image in the validation set is classified and given an uncertainty score. The top $k$ uncertain images are transferred from the validation set to the training set and the process repeats for $M$ active learning iterations. Finally, we select the best performing model at the $m^{th}$ iteration based on the F1 score when evaluated on the hold-out test set. \textbf{Fig. 2} provides an overview of a single active learning iteration.

It is important to consider that OCTA500 contains 304 images per subject, which means transferring instances on an individual image basis could result an an upward bias for classes in which images belonging to the same subject appear in both the training and validation set. As such, we also implement subject-based sampling to avoid data leakage. This sampling method has a two phase approach, where phase 1 averages the uncertainty scores for each class to determine the overall most difficult class to identify, then in phase 2, we repeat the process for all subjects belonging to the selected class. Finally, all images associated with the $k$ most difficult subjects are transferred to the train set. 

\begin{figure}[t]
    \centering
    \includegraphics[width=\textwidth]{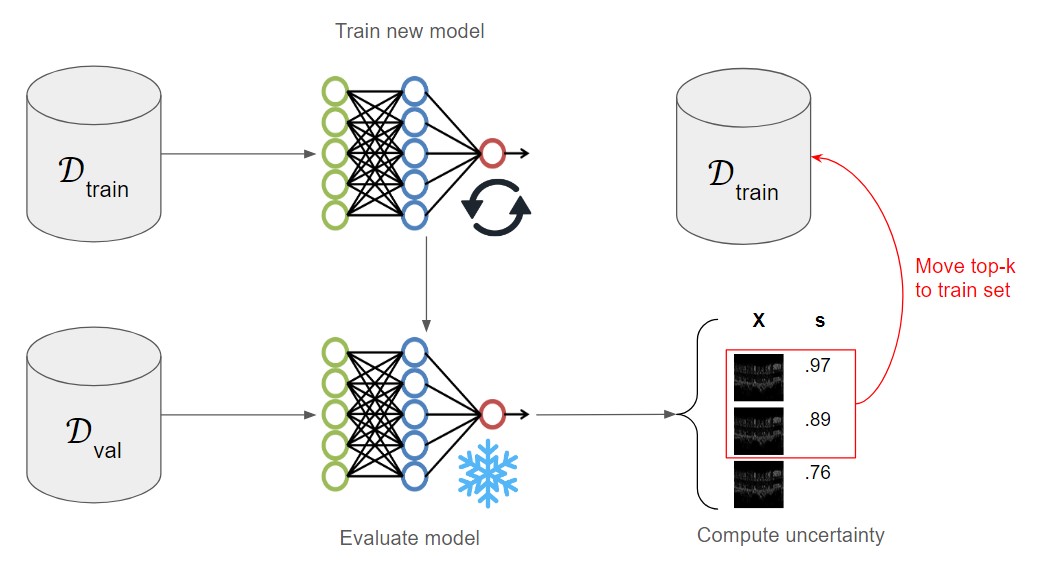}
    \caption{Example of a single active learning iteration, where $\mathcal{D}_{\{train, val\}}$ are the train and validation set, $s$ is the uncertainty score, and $k$ is the number of images to move at each iteration}
    \label{fig:AL}
\end{figure}

\section{Experiments}

\subsection{Baselines}
We evaluate our approach against conventional methods for handling highly imbalanced datasets. We first attempt to work with the imbalanced data directly by applying Inverse Frequency Class Weighting to each of the classes during Cross Entropy loss calculation. Here, each class is weighted inversely proportionate to the amount of instances appearing in the train dataset. This method aims to mitigate class bias by weighing minority classes more heavily than the over represented class. 

We then consider methods for balancing the data. We first randomly undersample the majority classes (Normal, DR) such that the dataset is perfectly balanced, with four subjects representing each class. Then, we experiment with oversampling the minority classes (AMD, CNV) by doubling and tripling the number of appearances in each epoch (Note: The results demonstrated in \textbf{Table 1} represent the experiments with doubled instances). We additionally add data augmentation in the form of AutoAugment \cite{AutoAug} and AugMix \cite{hendrycks2020augmixsimpledataprocessing} to create variety in the oversampled classes.

\subsection{Active learning sampling}
We perform active learning experiments based on four uncertainty metrics. Let $s$ be the uncertainty score, where higher values indicate higher uncertainty (low confidence) and let $P(\hat{y}|X)$ be the set of Softmax probabilities corresponding to the predicted class $\hat{y}$ for each image in the batch $X$ as generated by the model $F(X)$. 

Least Confidence ($\mathcal{L}$) \cite{LC} uncertainty simply takes the Softmax probability of the predicted label as a measure of confidence. Here, a higher value indicates higher confidence, so we invert this value by subtracting it from 1 to get the uncertainty score:

\begin{equation}
    s_{\mathcal{L}} = \underset{X}{\text{argmax}}[1 - P(\hat{y}|X)]
\end{equation}

Margin sampling ($\mathcal{M}$) \cite{margin} instead considers the difference in certainty between the top two most confident predictions. A large difference between the top two predictions indicates a highly confident model, while a small difference suggests the image was difficult to classify. Again, we invert $\mathcal{M}$ to obtain an uncertainty score that matches our definition.

\begin{equation}
    s_{\mathcal{M}} = \underset{X}{\text{argmax}}[1 - (P(\hat{y}|X) - P(\hat{y}'|X))]
\end{equation}

Where $\hat{y}'$ is the second most confident predicted class. Similarly, we perform Ratio Sampling ($\mathcal{R}$) \cite{monarch_2021_humanintheloop}, which instead computes the confidence ratio between the top two predicted classes. Since the ratio has no upper bound (and is lower bounded by 1), where a large difference in confidence yields a higher number, we instead take the negative as the uncertainty score.

\begin{equation}
    s_{\mathcal{R}} = \underset{X}{\text{argmax}}[-\frac{(P\hat{y}|X)}{P(\hat{y}'|X)}]
\end{equation}

Finally, Entropy Sampling ($\mathcal{E}$) \cite{4563068} calculates the entropy of the predicted probability distribution as the uncertainty score via \textbf{Eq. 4}. Unlike the other forms of uncertainty, entropy considers the entire posterior distribution, rather than just the top one or two. This gives a more holistic view of the model's understanding of each class. 

\begin{equation}
    s_{\mathcal{E}} = \underset{X}{\text{argmax}}[-\sum_{i} P(\hat{y}_i|X)\log P(\hat{y}_i|X)]
\end{equation}

\subsection{Implementation details}
We utilized Inception-V3 \cite{szegedy2015rethinking} as the backbone model and replaced the output head with two linear layers with 512 and 4 nodes, respectively. We finetuned this model with the Adam optimizer and a learning rate of $2\times10^{-5}$ for 5 epochs. Active learning was applied for 10 iterations with $k=1$ for the subject-based method and $k=300$ for instance-based. The final model was selected from the AL iteration based on the highest F1 score.

\section{Results}
The main results in \textbf{Table 1} demonstrate a nearly 50\% and 49\% gain in accuracy and F1, respectively, for the top performing active learning model (ratio sampling), compared to the unbalanced baseline. Even entropy sampling, the worst performing AL model, saw considerable improvement over unbalanced, achieving a roughly 40\% and 27\% gain in accuracy and F1. Interestingly, each of the non-AL strategies saw a slight decrease in performance compared to the unbalanced baseline on OCTA500. The decrease in performance for the augmented oversampling methods suggest that augmentation techniques such as AutoAugment and AugMix are not suitable for OCTA data. Additionally, the performance of random undersampling demonstrates a strong need for active learning based undersampling, especially when compared to the performance of the AL strategies. It should be noted that since the performance of the AL experiments is substantially better than the baselines, we did not feel it was necessary to take an average over multiple trials as the variability created by randomness would not ultimately alter the results. As such, we provide results from a single, set-seed experiment.

\begin{table}[!t]

\begin{center}
\begin{tabular}{@{}lll@{}}
\toprule
Training method                                     & Acc      & F1    \\ \midrule
Unbalanced                                          & .5139    & .4864 \\
Inverse Frequency Class Weighting                   & .4956    & .4571 \\ \midrule
Random Undersampling                                & .4482    & .3334 \\
Oversampling (AutoAugment)                          & .4178    & .4136    \\ 
Oversampling  (AugMix)                              & .4647    & .4503    \\ \midrule
Least Confident Sampling                            & .7313    & .6285 \\
Entropy Sampling                                    & .7188    & .6187 \\
Margin Sampling                                     & .7282    & .6262 \\
\textbf{Ratio Sampling}                             & \textbf{.7688} & \textbf{.7116} \\ \bottomrule

\end{tabular}
\caption{Results from subject-based sampling experiments, where $k=1$}
\label{tab:my-table}
\end{center}
\end{table}

\subsection{Ablation analysis}
We additionally conduct an analysis of the performance of the standard instance-based sampling compared to the subject sampling. \textbf{Table 2} shows that subject sampling consistently outperforms the instance-based method. This falls in line with our intuition, as only moving a fraction of a subject's OCTA scans to the train dataset results in data leakage. The model then becomes overconfident with its predictions on the leaked instances, skewing the predictions, and thus the active learning pipeline.

\textbf{Table 2} also contains an ablation study of the hyperparameter $k$, which is responsible for determining how many elements to move to the train dataset after each AL iteration. For subject sampling, we chose $k \in \{1, 2\}$, and for instance sampling, we selected $k \in \{300, 600\}$ (approximately equal to the number of scans associated with 1 and 2 subjects, respectively). We found that $k$ did not have much impact on the performance for instance sampling, but subject sampling generally performs better with a lower $k$, with Least Confident performing slightly better at $k=2$.

\begin{table}[]
\begin{center}
\begin{tabular}{@{}lcccccccc@{}}
\toprule
                & \multicolumn{4}{c}{Instance sampling}                                   & \multicolumn{4}{c}{Subject Sampling}                         \\ \midrule
                & \multicolumn{2}{c|}{k=300}         & \multicolumn{2}{c|}{k=600}         & \multicolumn{2}{c|}{k=1}           & \multicolumn{2}{c}{k=2} \\ 
                & Acc   & \multicolumn{1}{c|}{F1}    & Acc   & \multicolumn{1}{c|}{F1}    & Acc   & \multicolumn{1}{c|}{F1}    & Acc        & F1         \\ \midrule
Least Confident & .6073 & \multicolumn{1}{c|}{.5145} & .6195 & \multicolumn{1}{c|}{.5354} & .7313 & \multicolumn{1}{c|}{.6285} & \textbf{.7536}      & \textbf{.6796}      \\
Entropy         & .6416 & \multicolumn{1}{c|}{.5466} & .5781 & \multicolumn{1}{c|}{.4744} & \textbf{.7188} & \multicolumn{1}{c|}{\textbf{.6187}} & .6496      & .5677      \\
Ratio           & .5962 & \multicolumn{1}{c|}{.5007} & .6059 & \multicolumn{1}{c|}{.5114} & \textbf{.7688} & \multicolumn{1}{c|}{\textbf{.7116}} & .6724      & .6195      \\
Margin          & .5986 & \multicolumn{1}{c|}{.5073} & .5976 & \multicolumn{1}{c|}{.5101} & \textbf{.7282} & \multicolumn{1}{c|}{.6262} & .7196      & \textbf{.6784}      \\ \bottomrule
\end{tabular}
\caption{Evaluation of Instance vs Subject sampling and ablation analysis of $k$}
\label{tab:my-table}
\end{center}
\end{table}

Finally, it has been observed that probabilities outputted by the Softmax function $\sigma(\cdot)$ are not true estimates of the actual model confidence. As such, \cite{guo2017calibration} has proposed a calibration technique whereby the probability distribution given by $\sigma(F(X))$ is scaled according to a temperature parameter $T$. Since each of the four AL sampling methods begin with Softmax probabilities, we explored the affect of calibration on the model's performance. In our implementation, we compute $T$ by optimizing the negative log-likelihood with respect to the hold-out test set at the end of each active learning iteration and take the calibrated probability distribution as $\sigma(F(X)/T)$. The results in \textbf{Table 3} indicate that Softmax calibration generally reduces performance of Least Confident, Ratio, and Margin sampling, but yields significantly higher performance for Entropy sampling. We believe this is due to the fact that $\mathcal{L}$, $\mathcal{R}$, and $\mathcal{M}$ only utilize the top 1 or 2 probabilities, and likely do not benefit much from calibration. However, $\mathcal{E}$ takes a more holistic view of the entire probability distribution and thus gains a considerable amount of additional information from a well-calibrated probability distribution.

It should be noted that wider and deeper models generally achieve higher discriminatory power at the cost of calibration \cite{guo2017calibration}. As such, calibration may play a smaller role in smaller models such as ResNet18 \cite{resnet} or a bigger role in larger models such as VGG16 \cite{simonyan_2014_very}. 

\begin{table}[]
\begin{center}
    
\begin{tabular}{@{}lllll@{}}
\toprule
                & \multicolumn{2}{c}{Uncalibrated}                   & \multicolumn{2}{c}{Calibrated}                   \\
                & \multicolumn{1}{c}{Acc}  & \multicolumn{1}{c}{F1}  & \multicolumn{1}{c}{Acc} & \multicolumn{1}{c}{F1} \\ \midrule
Least Confident & \textbf{.7313}           & \textbf{.6285}          & .7165                   & .6111                   \\
Entropy         & .7188                    & .6187                   & \textbf{.7717}          & \textbf{.7318}         \\
Ratio           & \textbf{.7688}           & \textbf{.7116}          & .7065                   & .6115                   \\
Margin          & .7282                    & \textbf{.7116}          & \textbf{.7303}          & .6243                  \\ \bottomrule
\end{tabular}
\caption{Analysis of the effect of calibration}
\label{tab:my-table}
\end{center}

\end{table}

\section{Conclusion}
In this study, we demonstrate that the straightforward application of deep learning is not capable of detecting retinal diseases from OCTA images. This can be attributed to limited labeled training data, and furthermore, available datasets are highly imbalanced and require additional data engineering to achieve quality performance. To address this issue, we presented various active learning strategies and found that these strategies greatly outperform traditional methods such as inverse frequency class weighting, undersampling, and oversampling.
We conducted several ablation studies, including analyzing instance vs. subject sampling and the role of network calibration for active learning. We found that model calibration becomes substantially more important as the active learning method examines a larger proportion of the probability distribution (e.g., entropy sampling).
Future studies in this direction should focus on further enhancing and refining the presented active learning approach to improve generalization performance by considering diversity sampling and hybrid approaches. Additionally, future research should analyze other aspects, including explainability, for identifying imaging biomarkers in OCTA imaging.

\subsubsection{Acknowledgments:} This research was supported by West Virginia Higher
Education Policy Commission’s Research Challenge Grant Program 2023 and
DARPA/FIU AI-CRAFT grant.

\bibliography{export}

\begin{thebibliography}{10}

\bibitem{eye}
Eye conditions and diseases | national eye institute.

\bibitem{a2015_eye}
Eye health statistics - american academy of ophthalmology, 2015.

\bibitem{a2024_choroidal}
Choroidal neovascularization: Oct angiography findings - eyewiki, 05 2024.

\bibitem{AutoAug}
Ekin~Dogus Cubuk, Barret Zoph, Dandelion Man{\'{e}}, Vijay Vasudevan, and Quoc~V. Le.
\newblock Autoaugment: Learning augmentation policies from data.
\newblock {\em CoRR}, abs/1805.09501, 2018.

\bibitem{fricke_2018_global}
Timothy~R. Fricke, Nina Tahhan, Serge Resnikoff, Eric Papas, Anthea Burnett, Suit~May Ho, Thomas Naduvilath, and Kovin~S. Naidoo.
\newblock Global prevalence of presbyopia and vision impairment from uncorrected presbyopia.
\newblock {\em Ophthalmology}, 125:1492--1499, 10 2018.

\bibitem{fujimoto_2000_optical}
James~G. Fujimoto, Costas Pitris, Stephen~A. Boppart, and Mark~E. Brezinski.
\newblock Optical coherence tomography: An emerging technology for biomedical imaging and optical biopsy.
\newblock {\em Neoplasia}, 2:9--25, 01 2000.

\bibitem{guo2017calibration}
Chuan Guo, Geoff Pleiss, Yu~Sun, and Kilian~Q. Weinberger.
\newblock On calibration of modern neural networks, 2017.

\bibitem{resnet}
Kaiming He, Xiangyu Zhang, Shaoqing Ren, and Jian Sun.
\newblock Deep residual learning for image recognition.
\newblock {\em CoRR}, abs/1512.03385, 2015.

\bibitem{hendrycks2020augmixsimpledataprocessing}
Dan Hendrycks, Norman Mu, Ekin~D. Cubuk, Barret Zoph, Justin Gilmer, and Balaji Lakshminarayanan.
\newblock Augmix: A simple data processing method to improve robustness and uncertainty, 2020.

\bibitem{4563068}
Alex Holub, Pietro Perona, and Michael~C. Burl.
\newblock Entropy-based active learning for object recognition.
\newblock In {\em 2008 IEEE Computer Society Conference on Computer Vision and Pattern Recognition Workshops}, pages 1--8, 2008.

\bibitem{LC}
David~D. Lewis and William~A. Gale.
\newblock A sequential algorithm for training text classifiers.
\newblock {\em CoRR}, abs/cmp-lg/9407020, 1994.

\bibitem{li2024octa}
Mingchao Li, Kun Huang, Qiuzhuo Xu, Jiadong Yang, Yuhan Zhang, Zexuan Ji, Keren Xie, Songtao Yuan, Qinghuai Liu, and Qiang Chen.
\newblock Octa-500: a retinal dataset for optical coherence tomography angiography study.
\newblock {\em Medical Image Analysis}, 93:103092, 2024.

\bibitem{monarch_2021_humanintheloop}
Robert Monarch and Christopher~D Manning.
\newblock {\em Human-in-the-loop machine learning : active learning and annotation for human-centered AI}.
\newblock Shelter Island Manning, 2021.

\bibitem{rasti_2018_macular}
Reza Rasti, Hossein Rabbani, Alireza Mehridehnavi, and Fedra Hajizadeh.
\newblock Macular oct classification using a multi-scale convolutional neural network ensemble.
\newblock {\em IEEE Transactions on Medical Imaging}, 37:1024–1034, 04 2018.

\bibitem{margin}
Dan Roth and Kevin Small.
\newblock Margin-based active learning for structured output spaces.
\newblock In Johannes F{\"u}rnkranz, Tobias Scheffer, and Myra Spiliopoulou, editors, {\em Machine Learning: ECML 2006}, pages 413--424, Berlin, Heidelberg, 2006. Springer Berlin Heidelberg.

\bibitem{simonyan_2014_very}
Karen Simonyan and Andrew Zisserman.
\newblock Very deep convolutional networks for large-scale image recognition.
\newblock {\em Computer Science}, 2014.

\bibitem{sousa2020optical}
David~Cordeiro Sousa, In{\^e}s Leal, Susana Moreira, S{\'o}nia do~Vale, Ana~R Silva-Herdade, Patr{\'\i}cia Dion{\'\i}sio, Miguel~ARB Castanho, Lu{\'\i}s Abeg{\~a}o~Pinto, and Carlos Marques-Neves.
\newblock Optical coherence tomography angiography study of the retinal vascular plexuses in type 1 diabetes without retinopathy.
\newblock {\em Eye}, 34(2):307--311, 2020.

\bibitem{spaide2018optical}
Richard~F Spaide, James~G Fujimoto, Nadia~K Waheed, Srinivas~R Sadda, and Giovanni Staurenghi.
\newblock Optical coherence tomography angiography.
\newblock {\em Progress in retinal and eye research}, 64:1--55, 2018.

\bibitem{szegedy2015rethinking}
Christian Szegedy, Vincent Vanhoucke, Sergey Ioffe, Jonathon Shlens, and Zbigniew Wojna.
\newblock Rethinking the inception architecture for computer vision, 2015.

\bibitem{wang_2019_on}
Depeng Wang and Liejun Wang.
\newblock On oct image classification via deep learning.
\newblock {\em IEEE Photonics Journal}, 11:1--14, 10 2019.

\bibitem{zhou_2023_a}
Yukun Zhou, Mark~A. Chia, Siegfried~K. Wagner, Murat~S. Ayhan, Dominic~J. Williamson, Robbert~R. Struyven, Timing Liu, Moucheng Xu, Mateo~G. Lozano, Peter Woodward-Court, Yuka Kihara, Andre Altmann, Aaron~Y. Lee, Eric~J. Topol, Alastair~K. Denniston, Daniel~C. Alexander, and Pearse~A. Keane.
\newblock A foundation model for generalizable disease detection from retinal images.
\newblock {\em Nature}, 622:1–8, 09 2023.

\end{thebibliography}
\bibliographystyle{plain}
\end{document}


\section*{Supplementary Materials}

\subsection{DeepHit framework for time-to-event prediction}
For our task head, we adopt the DeepHit framework, a popular deep learning approach to survival analysis. With DeepHit framework, instead of predicting a single hazard coefficient for a given input, we output a distribution of hazards at discrete time points. This allows the model to learn the first hitting times (predicted time until the occurrence of the first event of interest for each subject in the study) directly without making assumptions about the underlying form of the data. 
In specific, the 
model learns to minimize the loss function $\mathcal{L}_{total} = \mathcal{L}_1 + \mathcal{L}_2$, where $\mathcal{L}_1$ is the log-likelihood of the distribution of the hitting time, defined as

\begin{equation}
    \mathcal{L}_1 = -\sum^{N}_{i=1}[\mathbbm{1}(\delta_i = 1) * \log{h_i^{T_i}} + \mathbbm{1}(\delta_i \neq 1) * \log{(1 - \hat{F}(T_i|x_i)}]
\end{equation}
where, $\mathbbm{1}$ is an indicator function evaluating to 1 iff $\delta_i = 1$ (event occured), $h_i^{T_i}$ corresponds to the predicted hazard for input $X_i$ at time $T_i$, and $\hat{F}(T_i|x_i)$ is the estimated cumulative incidence function (CIF) which approximates the probability that the event will occur on or before time $T_i$. $\mathcal{L}_2$ incorporates a combination of cause-specific
ranking loss functions and is defined as:
\begin{equation}
    \mathcal{L}_2 = \gamma \sum_{i \neq j} A_{i, j} \cdot \eta (\hat{F}(T_i | x_i), \hat{F}(T_i | x_j))
\end{equation}
where $\gamma$ is a hyperparameter which indicates the intensity of the ranking loss, $A_{i, j} = \mathbbm{1}(T_i < T_j)$ represents an inidcator function which evaluates to $1$ if a pair $(i, j)$ experience an event at different times, and $\eta(\cdot, \cdot)$ is a convex loss function, conveniently set as $\exp(\cdot)$ function.